%% file: main.tex
\documentclass{article}

\usepackage[preprint]{neurips_2026}
\usepackage{multirow}
\usepackage{pifont}
\usepackage[utf8]{inputenc} 
\usepackage[T1]{fontenc}    
\usepackage{hyperref}       
\usepackage{url}            
\usepackage{booktabs}       
\usepackage{amsfonts}       
\usepackage{nicefrac}       
\usepackage{microtype}      
\usepackage{xcolor} 
\usepackage{float}
\usepackage{pdfpages}
\usepackage{listings}
\usepackage{stackengine}
\usepackage{amsmath}
\usepackage{enumitem}
\usepackage{wrapfig}
\lstdefinelanguage{json}{
  basicstyle=\ttfamily\footnotesize,
  numbers=none,
  showstringspaces=false,
  breaklines=true,
  frame=single,
  rulecolor=\color{black!30},
  stringstyle=\color{purple!70!black},
}
\usepackage{graphicx}
\newcommand{\cmark}{\ding{51}} 
\newcommand{\xmark}{\ding{55}} 
\usepackage{graphicx}
\usepackage{booktabs}
\usepackage{multirow}
\usepackage[table]{xcolor}
\usepackage{tabularx}
\usepackage{array}
\usepackage{makecell}
\usepackage[most]{tcolorbox}
\usepackage{xcolor}
\usepackage{setspace}
\newcolumntype{L}[1]{>{\hsize=#1\hsize\raggedright\arraybackslash}X}
\newcolumntype{C}[1]{>{\hsize=#1\hsize\centering\arraybackslash}X}

\title{IPIBench: Evaluating Interactive Proactive Intelligence of MLLMs under Continuous Streams}

\author{
\bfseries
Jinzhao Li$^{1,2}$,
Yinuo Chen$^{1}$,
Wenxuan Song$^{1}$,
Yijia Lei$^{1}$,\\
\bfseries
Yichi Zhang$^{1}$,
Honglei Yan$^{2}$,
Panwang Pan$^{2}$,
Miao Liu$^{1\dag}$ \\[1em]
$^{1}$College of AI, Tsinghua University \\ $^{2}$ByteDance \\[1em]
\texttt{lijinzha22@mails.tsinghua.edu.cn} \quad \texttt{miaoliu@mail.tsinghua.edu.cn}
}

\begin{document}

\maketitle
\renewcommand{\thefootnote}{}
\footnotetext{$\dagger$ Corresponding author.}

\input{sec/0_abstract}
\input{sec/1_intro}
\input{sec/2_relatedwork}
\input{sec/3_datasets}
\input{sec/4_method}

\input{sec/5_experiments}
\input{sec/6_conclusion}

\bibliographystyle{abbrv}
\bibliography{main}

\end{document}

%% file: sec/0_abstract.tex
\definecolor{proactivepurple}{HTML}{C4B4FF}
\definecolor{proactiveblue}{HTML}{74D4FF}
\definecolor{proactiveindigo}{HTML}{A3B3FF}

\begin{abstract}
Recent multimodal large language models (MLLMs) achieve strong performance on reactive question answering, but real-world streaming assistants require proactive reasoning over continuous visual inputs. Existing benchmarks mainly study reactive or proactive interactions in isolated single-turn settings, overlooking dynamic multi-turn scenarios where users may add, modify, or cancel proactive requests alongside interleaved reactive queries. To address this gap, we introduce \textbf{IPIBench}, the first benchmark for evaluating Interactive Proactive Intelligence of MLLMs under streaming video settings. IPIBench covers proactive monitoring, proactive task management, and interleaved reactive–proactive requests. Evaluations on representative MLLMs reveal two major limitations: unstable proactive triggering and weak coordination between reactive and proactive behaviors. We further propose \textbf{IPI-Agent}, a training-free agentic framework with an interaction-control policy and a temporal-gating mechanism for stabilizing proactive triggering and coordinating multi-turn interactions. Experiments show that IPI-Agent consistently improves existing MLLMs across all benchmark settings. Project page: \textcolor{magenta}{\url{https://lijinzhao30.github.io/IPIBench/}}
\end{abstract}

\begin{figure}[t]
    \centering
    \includegraphics[width=\linewidth]{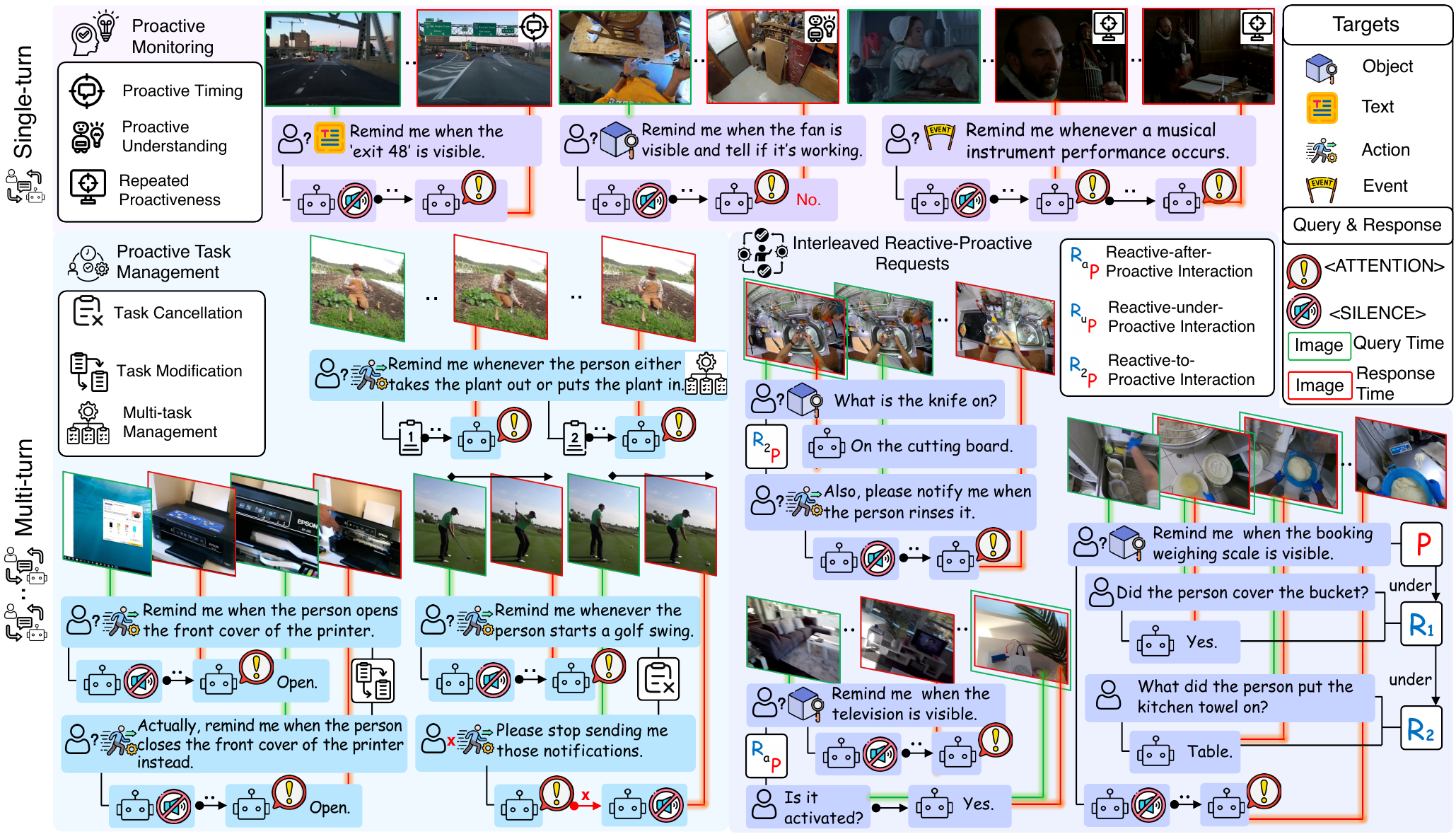}
    \vspace{-2.0em}
    \caption{\textit{Visual examples of our proposed IPI-Bench}. For single-turn \textcolor{proactivepurple}{proactive monitoring} tasks, our benchmark covers timing, understanding, and repeated proactiveness. For multi-turn interactive proactive tasks, the benchmark further addresses \textcolor{proactiveblue}{proactive task management} (e.g., modification, cancellation, and multi-task management) and \textcolor{proactiveindigo}{interleaved reactive–proactive requests}.}
    \vspace{-1.5em}
    \label{fig:teaser}
\end{figure}

%% file: sec/1_intro.tex
\section{Introduction}

Existing multimodal large language models (MLLMs)\cite{liu2023visualinstructiontuning}\cite{hong2025glm}\cite{hong2025glm}\cite{singh2025openai}excel at reactive question answering, benefiting from instruction tuning that optimizes models to respond to explicit user queries. Alternatively, the emergence of always-on platforms, such as wearable AI devices\cite{wen2025aiserviceproactiveassistance} and home assistant robots\cite{driess2023palmeembodiedmultimodallanguage}\cite{Wu_2023}, demands models capable of providing timely assistance by proactively reasoning over continuous perceptual inputs. Despite a handful of recent efforts\cite{xu2025streamingvlmrealtimeunderstandinginfinite}\cite{yang2025streammemqueryagnostickvcache}on streaming MLLMs, proactive capabilities are still largely studied in fixed single-turn settings without subsequent adaptation or follow-up queries.

However, in real-world scenarios, users may constantly adjust ongoing proactive interactions or introduce new reactive queries. We provide visual examples in Fig.~1, showcasing multi-turn settings where users can add, edit, or cancel previous proactive queries alongside interleaved proactive-reactive interactions. 
Therefore, the primary goal of this work is to investigate the modeling of proactive requests in dynamic and interactive settings.


Existing streaming video understanding benchmarks typically study reactive and proactive interactions in isolation through single-turn VQA formulations. In contrast, we introduce \textbf{IPIBench}, the first benchmark for evaluating \textbf{I}nteractive \textbf{P}roactive \textbf{I}ntelligence of MLLMs. 
Our benchmark begins with single-turn tasks covering timing, understanding, and repeated proactiveness. We then extend to multi-turn scenarios, including follow-up proactive management operations such as add, edit, and cancel, as well as different forms of interleaved proactive-reactive requests based on their relations and temporal order. A comparison between our benchmark and prior work is provided in Tab.~\ref{tab:relatedwork}


We evaluate prevailing MLLMs and online VLMs on IPIBench. Interestingly, we find that existing models fail substantially in this challenging setting due to the lack of a unified interaction policy, leading to missed events, premature or delayed responses, and difficulty in coordinating reactive and proactive behaviors under continuous streams. To address these limitations, we propose \textbf{IPI-Agent}, a training-free agentic framework that introduces an \emph{Interaction-Control Policy} for coordinating reactive queries, proactive instructions, and management instructions, together with a \emph{Temporal-Gating Mechanism} for stabilizing proactive triggering. Experiments show that IPI-Agent consistently improves base MLLMs across all tasks.
Our key contributions are summarized as follows:
\begin{itemize}[nosep,leftmargin=10pt]
    \item We introduce IPIBench, the first benchmark for evaluating interactive proactive intelligence of MLLMs under streaming video settings, covering proactive monitoring, proactive task management, and interleaved reactive--proactive requests.
    \vspace{0.2em}
    \item We conduct systematic evaluations and failure analyses on representative proprietary, open-source, and online streaming models, revealing two key limitations of existing MLLMs under interactive streaming settings: unstable proactive triggering and weak multi-turn interaction coordination.
    \vspace{0.2em}
    \item We propose IPI-Agent, a training-free agentic framework with an interaction-control policy and a temporal-gating mechanism, which consistently improves proactive triggering stability and multi-turn interaction coordination.
\end{itemize}

%% file: sec/2_relatedwork.tex
\section{Related Work}

\subsection{Streaming Video Benchmarks}

Existing streaming video benchmarks mainly evaluate causal understanding over online video inputs. OVBench \cite{huang2025online}, SVBench \cite{yang2025svbench}, StreamBench \cite{xiong2025streaming}, and RTV-Bench \cite{xunrtv} study real-time perception, temporal  memory, reactive interaction, and reasoning over evolving contexts. TemporalBench \cite{cai2024temporalbench} and StreamingCoT \cite{hu2025streamingcot} further emphasize fine-grained temporal reasoning and temporally evolving rationales. More recently, RIVER-Bench \cite{shiriver} evaluates retrospective memory, live perception, and streaming narration under online video settings. These benchmarks demonstrate that MLLMs struggle with partial and continuously evolving evidence under streaming inputs. However, their evaluation protocols are still largely formulated as reactive question answering.

Recent proactive benchmarks further investigate whether models can respond proactively at appropriate moments.  StreamingBench \cite{lin2026streamingbench}, OVO-Bench \cite{niu2025ovo}, PROASSIST \cite{zhang2025proactive}, ProactiveVideoQA \cite{wang2025proactivevideoqa}, ProReady-QA \cite{azad2026streamready}, and ESTP-Bench \cite{zhang2025eyes} evaluate proactive triggering, live assistance, and evidence-ready answering. MMDuet \cite{wang2024videollm} and OmniMMI \cite{wang2025omnimmi} further extend evaluation to grounded response insertion and multimodal interaction. These benchmarks study proactive behavior in fixed single-turn settings or under isolated evaluation protocols, and often lack systematic and fine-grained evaluation of proactive triggering behavior. In contrast, IPIBench focuses on interactive proactive intelligence of MLLMs under sustained streaming scenarios, where proactive monitoring, task management, and reactive interactions dynamically coexist over time.

\begin{table}[t]
    \centering
    \caption{\textbf{Comparison of representative streaming video benchmarks.} We compare benchmarks by interaction pattern, multi-turn setting, visual source, task form, number of QA pairs, and annotation source. Our benchmark targets interactive proactive intelligence, where proactive monitoring, task management, and reactive interactions coexist in continuous video streams.}
    \vspace{-0.5em}
    \label{tab:relatedwork}
    \resizebox{\linewidth}{!}{
        \begin{tabular}{@{}lcccccc@{}}
        \toprule
        \textbf{Benchmark} & \textbf{Interaction Pattern} &  \textbf{Multi-turn} & \textbf{Visual Source} & \textbf{Task Form} & \textbf{QA Pairs} & \textbf{Annotation} \\ 
        \midrule
        OVO-Bench \cite{niu2025ovo} & Reactive or Proactive & \xmark & Both & MCQ, Open  &  2,814  & MLLM+Human \\
        RTV-Bench \cite{xunrtv} & Reactive & \xmark & Both  & MCQ  & 4,608  & LLM+Human\\
        StreamingBench \cite{lin2026streamingbench} & Reactive or Proactive & \xmark & Both & Open & 4,500 &  MLLM+Human\\
        ESTP-Bench \cite{zhang2025eyes} & Proactive & \xmark & Ego. & Open & 2,264 &  LLM+Human\\
        ProactiveVideoQA \cite{wang2025proactivevideoqa} & Proactive & \xmark & Both  &  Open & 1,427 & LLM+Human \\
        StreamingCoT \cite{hu2025streamingcot} & Reactive & \xmark & Exo. & MCQ & 34,470 & MLLM + Human \\
        RIVER Bench \cite{shiriver} & Reactive or Proactive &\xmark & Both & MCQ, Open & 4,278 &  LLM+Human\\
        \rowcolor{gray!15} Ours & Interactive Proactive & \cmark & Both & Open & 3,738 & LLM+Human \\
        \bottomrule
        \end{tabular}
    }
\label{tab:relatedwork}
\end{table}

\subsection{Proactive Streaming Video Models}
Streaming video models move from offline comprehension to incremental perception and response. VideoLLM-online \cite{chen2024videollm} introduces streaming EOS prediction to decide whether the model should respond or remain silent. VideoLLM-MoD \cite{wu2024videollm} improves efficiency with mixture of depths vision computation, and LION-FS \cite{li2025lion} separates fast timing decisions from slower detailed reasoning. Streamo \cite{xia2025streaming} broadens task coverage through large scale streaming instruction tuning. These models make online video dialogue more practical, but they typically treat prompt conditioned answering and autonomous response emission as separate behaviors.

A complementary line makes response timing explicit. MMDuet \cite{wang2024videollm} formulates video text duet interaction for deciding when to speak during playback, and MMDuet2 \cite{wang2025mmduet2} introduces no reply actions and multi turn reinforcement learning. StreamReady \cite{azad2026streamready} studies readiness aware answering once sufficient evidence appears, while EgoSpeak \cite{kim2025egospeak} focuses on real time speech initiation from egocentric video. StreamMind \cite{ding2025streammind} invokes cognition at eventful moments, Dispider \cite{qian2025dispider} decouples perception, decision, and reaction modules, and StreamAgent \cite{yang2025streamagent} uses anticipatory planning to guide streaming agents. These works clarify when a system should observe, wait, or speak, but their policies are usually optimized around a standing query, a standing goal, or event triggered speaking. In contrast, our setting requires models to continuously coordinate proactive monitoring, task management, and interleaved reactive-proactive requests under evolving streaming contexts.

\subsection{Video Large Language Models}

Recent video large language models have substantially improved multimodal perception and reasoning through instruction tuning and unified image--video modeling. Representative works include Video-ChatGPT \cite{maaz2024video}, InternVL \cite{chen2024internvl}, VideoGPT+ \cite{maaz2024videogpt+}, Oryx \cite{liuoryx}, Qwen2-VL \cite{wang2024qwen2}, LLaVA-OneVision \cite{li2024llava}, Qwen2.5-VL \cite{Bai2025Qwen25VLTR}, and Qwen3-VL \cite{bai2025qwen3}. Long-video models such as LongVA \cite{zhang2024long}, InternVideo2.5 \cite{wang2025internvideo2}, Keye-VL-1.5 \cite{yang2025kwai}, and SlowFast-LLaVA-1.5 \cite{xu2025slowfast} further improve temporal coverage and long-context modeling. Despite these advances, existing Video-LLMs remain primarily optimized for reactive interactions, lacking unified mechanisms for proactive monitoring and multi-turn interaction coordination under continuous streams.

%% file: sec/3_datasets.tex
\section{IPIBench}
\label{sec:benchmark}

We now present \textbf{IPIBench}, a benchmark for evaluating interactive proactive intelligence in streaming video settings. In following sections, we first introduce the task definition and taxonomy. We then describe the construction process of the benchmark. Finally, we provide statistical analyses to illustrate the diversity and characteristics of the dataset.

\subsection{Task Definition and Taxonomy}

\label{sec:taxonomy}


We formalize the definition of proactive tasks with evolving user interactions. At each time step $t$, the model observes a video sequence $\mathbf{x}_{\le t} = \{x_1, x_2, \dots, x_t\}$ together with an optional user query $u$ regarding a future event. During the streaming process, these tasks may be dynamically updated through management instructions, e.g. cancelling or editing tasks. Meanwhile, users may also issue reactive queries that require immediate responses based on the current or historical visual context.


Concretely, the model maintains a set of active proactive tasks $\mathcal{T}_t = \{\tau_1, \tau_2, \dots, \tau_N\}$ at time $t$, where each task $\tau_i$ defines a target condition over the observed video stream. At each time step, the model produces an output $y_t$, which may correspond to a proactive trigger, a task update, an immediate response or no action. Importantly, under the streaming setting, the model only has access to frames up to time $t$, and must make decisions without future information. The objective is to correctly determine when to respond, what to respond, and how to update $\mathcal{T}_t$ under continuous streams. We therefore consider three interaction types: \textbf{Proactive Monitoring}, \textbf{Proactive Task Management}, and \textbf{Interleaved Reactive-Proactive Requests}


\noindent\textbf{Proactive Monitoring}.\ This aspect focuses on single-turn proactive tasks that evaluate whether the model can respond proactively at appropriate moments based on $\mathbf{x}_{\le t}$:

\begin{itemize}[nosep,leftmargin=10pt]
    \item \emph{Proactive Timing}: This task evaluates temporal precision by requiring the model to trigger a response when a specified event occurs. 
    \vspace{0.1em}
    
    \item \emph{Proactive Understanding}: This task requires the model to trigger at the correct time, while producing correct responses, such as describing attributes, spatial relations, or object states. 
    \vspace{0.1em}
    
    
    \item \emph{Repeated Proactiveness}: This task evaluates whether the model can consistently trigger responses for recurring events over time, rather than detecting only a single occurrence.
\end{itemize}




\noindent\textbf{Proactive Task Management}.\ This aspect evaluates user interactions that manage proactive tasks.
\begin{itemize}[nosep,leftmargin=10pt]
\item \emph{Task Cancellation}: Aproactive instruction is first issued to define a monitoring objective. In a subsequent turn, the user provides a cancellation instruction. The model is required to correctly cancel the task and stop responding to its corresponding target thereafter. 
\vspace{0.1em}
\item \emph{Task Modification}: A proactive instruction is first issued to define a monitoring objective. In a subsequent turn, the user provides a modification instruction. The model is required to correctly update the task according to the new instruction, and trigger responses based on the updated target. 
\item \emph{Multi-task Management}: The user issues an instruction that specifies multiple targets to monitor simultaneously. The model is required to correctly interpret the instruction and respond whenever any of the specified targets appears in the subsequent video stream.
\end{itemize}

\noindent\textbf{Interleaved Reactive-Proactive Requests}.\ This aspect evaluates complex multi-turn scenarios where reactive queries and proactive tasks are interleaved over time under the streaming setting. 
\begin{itemize}[nosep,leftmargin=10pt]
\item \emph{Reactive-after-Proactive Interaction}: This task considers scenarios where a reactive query is issued immediately after a proactive trigger, requiring the model to ground its response in both the triggering event and recent visual context.
\vspace{0.1em}

\item \emph{Reactive-under-Proactive Interaction}: This task considers cases where a reactive query is issued while a proactive task remains active, requiring the model to answer the reactive query while continuing to monitor the proactive task.
\vspace{0.1em}

\item \emph{Reactive-to-Proactive Interaction}: This task evaluates transitions from reactive to proactive interactions, where the model must correctly initialize and track new tasks based on user intent.
\end{itemize}

\subsection{Benchmark Construction}
\label{sec:construction}

We construct IPIBench in a progressive manner, starting from single-turn proactive monitoring tasks and extending them into increasingly complex interactive streaming scenarios. We first collect videos with temporal interval annotations of objects, scene text, actions, and events from diverse public datasets under both egocentric and exocentric settings, including Ego4D~\cite{grauman2022ego4d}, EgoTracks~\cite{tang2023egotracks}, QA-Ego4D~\cite{barmann2022did}, RoadTextVQA~\cite{tom2023reading}, COIN~\cite{tang2019coin}, Charades-STA~\cite{gao2017tall}, Oops~\cite{epstein2020oops}, QVHighlights~\cite{lei2021detecting}, AVA~\cite{gu2018ava}, YouCook2~\cite{zhou2018towards}, and THUMOS14~\cite{idrees2017thumos}. Based on these temporally grounded annotations, we first construct single-turn \emph{Proactive Monitoring} tasks by refining trigger moments under streaming settings through human annotation. We then progressively extend these instances into \emph{Proactive Task Management} and \emph{Interleaved Reactive--Proactive Requests} by inserting management instructions and reactive queries at different temporal positions within the video stream. Finally, we apply both automatic filtering and human verification to ensure accurate temporal grounding, natural interaction flow, and consistency with streaming constraints without future information leakage.

\subsection{Benchmark Statistics}
\label{sec:statistics}

\begin{figure}[t]
    \centering
    \includegraphics[width=\linewidth]{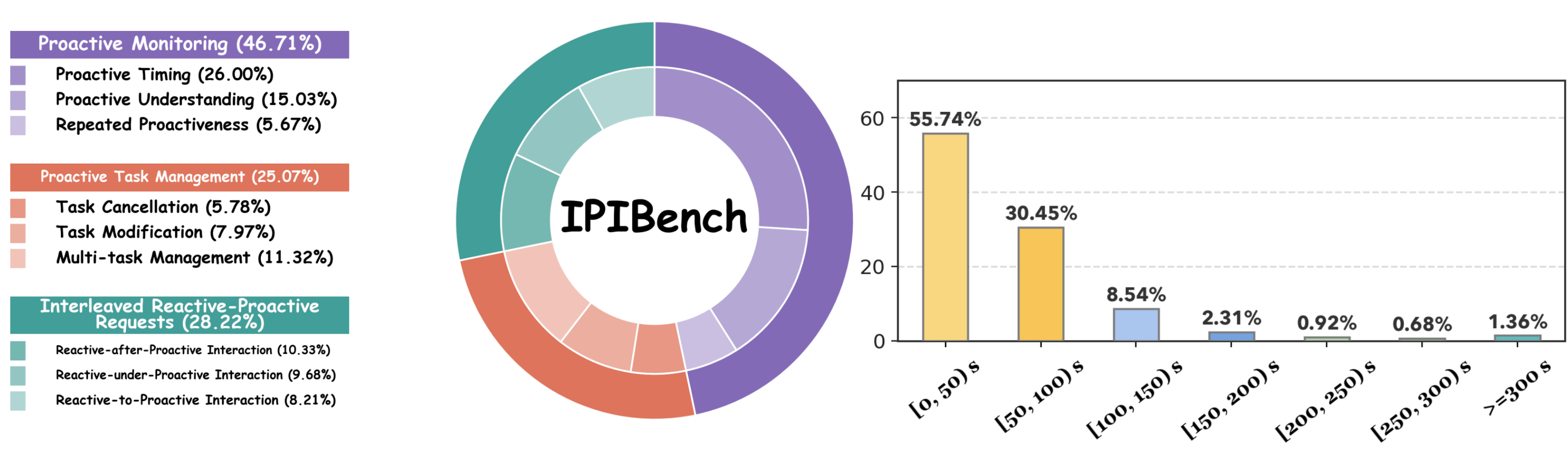}
    \caption{Statistics of our IPIBench. Left: Distribution of the three formulated task categories and subcategories. 
    Right: Distribution of video durations used for benchmark construction. }
    \label{fig:bench}
\end{figure}

We present the statistics of IPIBench in Fig.~\ref{fig:bench}. In total, IPIBench contains 1,831 videos and 3,738 QA instances, providing a comprehensive testbed for evaluating interactive proactive intelligence under streaming settings. As shown in Fig.~\ref{fig:bench} (left), the benchmark covers diverse task categories spanning proactive monitoring, proactive task management, and interleaved reactive--proactive requests, including both single-turn and multi-turn interaction scenarios. The video durations shown in Fig.~\ref{fig:bench} (right) range from short clips at the second level to long videos exceeding five minutes, covering diverse temporal scales under continuous streaming settings. IPIBench is constructed from both egocentric and exocentric video sources across diverse domains, including daily activities, instructional videos, driving scenarios, sports, and movie clips. 

%% file: sec/4_method.tex
\definecolor{policy_purple}{RGB}{165, 94, 234} 
\definecolor{gating_green}{RGB}{38, 222, 129}  
\definecolor{tool_red}{RGB}{252, 92, 101}      
\definecolor{tool_blue}{RGB}{69, 170, 242}     
\definecolor{tool_orange}{RGB}{250, 152, 58}   

\begin{figure}[t]
    \centering
    \includegraphics[width=\linewidth]{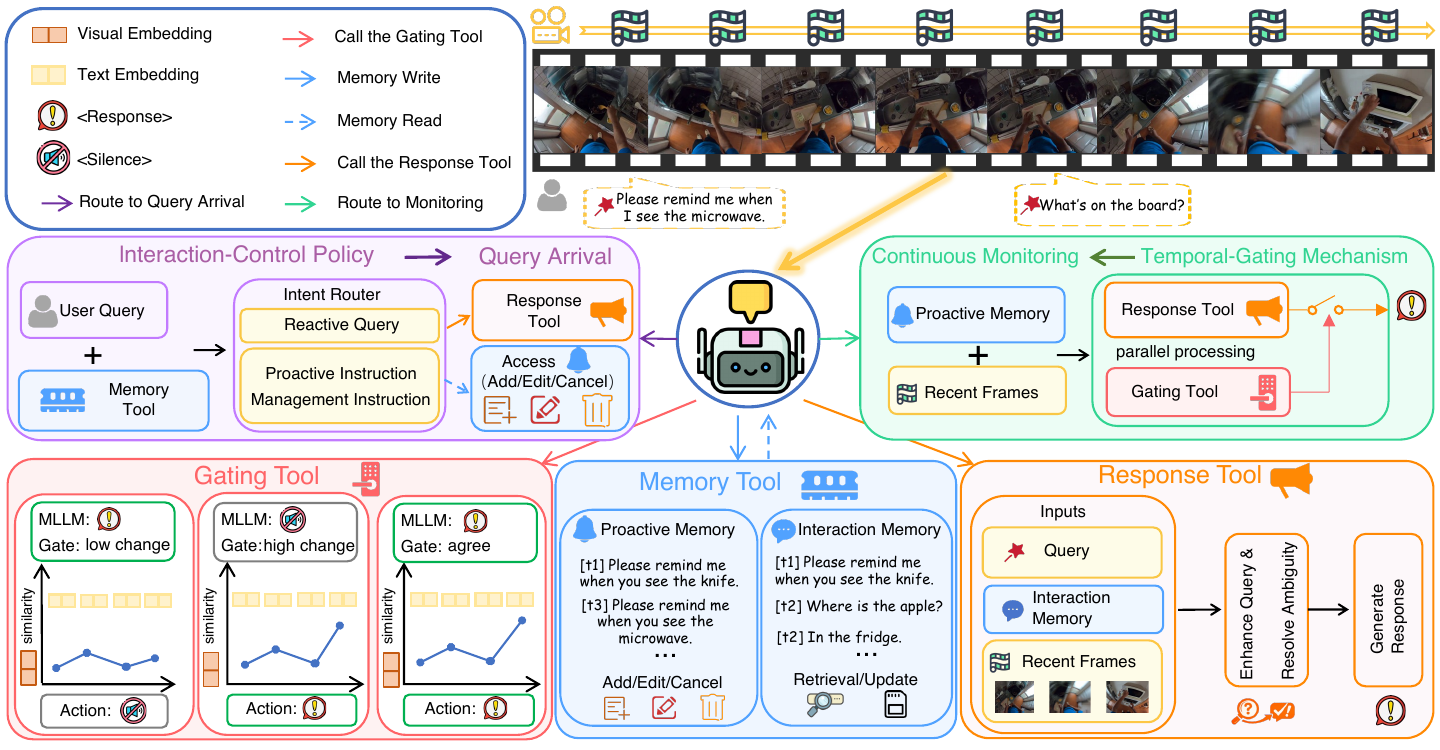}
    \vspace{-2.0em}
    \caption{\emph{Overview of IPI-Agent, a training-free agentic framework for interactive proactive behavior under streaming settings.} IPI-Agent operates under two workflows: Query Arrival and Continuous Monitoring. The framework maintains a unified \textcolor{policy_purple}{Interaction-Control Policy} through a \textcolor{tool_blue}{Memory Tool} and an Intent Router to coordinate reactive, proactive, and management instructions. In addition, IPI-Agent introduces a \textcolor{gating_green}{Temporal-Gating Mechanism} through a \textcolor{tool_orange}{Response Tool} and a \textcolor{tool_red}{Gating Tool} to improve proactive triggering stability under continuous streams.}
    \label{fig:agent}
\end{figure}

\section{IPI-Agent}
\label{sec:method}

Existing MLLMs exhibit two major limitations in interactive streaming settings. First, models struggle with temporally precise proactive triggering even in single-turn proactive tasks, often producing premature or delayed responses under continuous streams. Second, models perform poorly in multi-turn interactive proactive tasks due to the lack of a unified interaction policy for coordinating reactive and proactive behaviors over time. To address these challenges, we propose \textbf{IPI-Agent}, a training-free agentic framework for interactive proactive behavior under streaming settings.

\subsection{Overview}
\label{sec:overview}
Fig.~\ref{fig:agent} illustrates the overall interaction workflow of IPI-Agent. IPI-Agent operates under two execution workflows according to the incoming stream: \textbf{Query Arrival} and \textbf{Continuous Monitoring}. Our key insight is to design an \textbf{Interaction-Control Policy} through intent routing and a \emph{memory tool}, enabling the system to coordinate reactive queries with proactive and management instructions. In addition, we introduce a \textbf{Temporal-Gating Mechanism} using a \emph{gating tool} to improve proactive triggering stability from the \emph{response tool}.  We describe these two mechanisms as well as the tools in our framework in the following sections .


\subsection{Interaction-Control Policy}
\label{sec:controller}

IPI-Agent maintains a unified \emph{Interaction-Control Policy} under continuous streams through a Memory Tool that manages proactive and interaction memory. Specifically, the Memory Tool contains a Proactive Memory for proactive monitoring and task management, and an Interaction Memory for storing historical interactions to enhance current user queries with contextual information. The Intent Router then classifies query interaction types, including: 
\vspace{-0.5em}
\begin{itemize}[nosep,leftmargin=10pt]
\item For reactive queries, IPI-Agent retrieves relevant interaction history from the Interaction Memory and invokes the Response Tool using recent frames to generate an immediate response.

\item For proactive instructions, IPI-Agent converts the instruction into a structured proactive task and stores it in the Proactive Memory for continuous monitoring.

\item For management instructions, IPI-Agent accesses the Proactive Memory and updates the corresponding proactive task through modify or cancel operations.
\end{itemize}
This interaction-control policy allows proactive objectives to persist throughout long streaming interactions while remaining compatible with dynamically interleaved reactive queries.

\subsection{Temporal-Gating Mechanism}
\label{sec:gating}

IPI-Agent implements the \emph{Temporal-Gating Mechanism} through the collaboration between the Gating Tool and the Response Tool to improve proactive triggering stability under continuous streams. During streaming inference, the Response Tool first enhances task instructions using the Interaction Memory to resolve ambiguous references, and then generates proactive responses conditioned on recent video frames. The Gating Tool further regulates whether these proactive responses should be activated according to temporal similarity variations between proactive task proposals and recent visual observations.

At time step $t$, the framework maintains active proactive tasks $\mathcal{T}_t = \{\tau_1, \tau_2, \dots, \tau_N\}$. Whenever a new proactive instruction is added into the Proactive Memory, the agent automatically generates $M$ candidate textual proposals $\mathcal{P}_i = \{p_i^1, p_i^2, \dots, p_i^M\}$ for the corresponding task $\tau_i$. Using an embedding model, we compute proposal embeddings $\mathbf{e}_i^m = E(p_i^m)$ and the visual embedding of recent sliding-window frames $\mathbf{v}_t = E(\mathbf{x}_{t-K:t})$. For each proposal, we compute the similarity score $s_{i,t}^m = \mathrm{sim}(\mathbf{e}_i^m, \mathbf{v}_t)$ and its temporal variation $\Delta s_{i,t}^m = s_{i,t}^m - s_{i,t-1}^m$. The final temporal change score for task $\tau_i$ is obtained by max pooling as $\Delta_i(t) = \max_m \Delta s_{i,t}^m$.

Different proactive tasks are processed independently and in parallel for efficient streaming inference. Let $r_i(t)\in\{0,1\}$ denote whether the Response Tool triggers task $\tau_i$ at time $t$. The Gating Tool regulates triggering using two thresholds. If $r_i(t)=1$ but $\Delta_i(t)<\theta_{\text{low}}$, the response is suppressed to avoid premature activation. Conversely, if $r_i(t)=0$ but $\Delta_i(t)>\theta_{\text{high}}$, the Gating Tool activates the corresponding proactive response to recover missed triggers. $\theta_{\text{low}}$ and $\theta_{\text{high}}$ are chosen empirically according to the capability of the base model. By explicitly modeling temporal similarity variations, the proposed mechanism suppresses unstable early triggering while recovering delayed responses under continuous streams.

%% file: sec/5_experiments.tex
\definecolor{oursgreen}{RGB}{220,245,220}
\newcolumntype{L}[1]{>{\hsize=#1\hsize\raggedright\arraybackslash}X}
\newcolumntype{C}[1]{>{\hsize=#1\hsize\centering\arraybackslash}X}

\definecolor{proactivepurple}{HTML}{C4B4FF}
\definecolor{proactiveblue}{HTML}{74D4FF}
\definecolor{proactiveindigo}{HTML}{A3B3FF}

\section{Experiments}
\label{sec:experiments}

\subsection{Evaluation Protocol}
\label{sec:protocol}
To ensure fair comparison, all models are evaluated under a unified streaming protocol with a frame rate of 1 FPS. For models without specialized memory mechanisms, we simulate online inference using a sliding window of the most recent 16 frames. 

For proactive tasks, we evaluate models within the interval $[t^\ast-4,\, t^\ast+4]$ around the ground-truth trigger time $t^\ast$, advancing the stream at 1-second intervals. A prediction is considered correct if the first proactive trigger falls within $[t^\ast-1,\, t^\ast+1]$. For Repeated Proactiveness and multi-turn interaction tasks, a prediction is considered correct only if all required triggers and interactions are correct. For reactive tasks, we adopt an open-ended evaluation protocol by comparing model outputs up to 20 semantically equivalent candidate answers using bidirectional substring matching.

\begin{table*}[t]
    \centering
    \scriptsize
    \caption{\textbf{Main results on IPIBench.} We evaluate representative proprietary and open-source models across \textit{Proactive Monitoring}, \textit{Proactive Task Management}, and \textit{Interleaved Reactive--Proactive Requests} tasks. The scores represent the performance metrics (\%) for each sub-category. The best performance among non-human models in each column is highlighted in \colorbox{red!20}{light red}.}
    \label{tab:main}
    \setlength{\tabcolsep}{2.2pt}
    \renewcommand{\arraystretch}{1.0}
    \begin{tabularx}{0.98\linewidth}{@{} L{2.55} *{12}{C{0.75}} @{}}
    \toprule
    \multirow{2}{*}{Model} 
    & \multicolumn{4}{c}{\cellcolor{proactivepurple!55}Proactive Monitoring} 
    & \multicolumn{4}{c}{\cellcolor{proactiveblue!45}Proactive Task Management} 
    & \multicolumn{4}{c@{}}{\cellcolor{proactiveindigo!55}Interleaved Reactive--Proactive} \\
    \cmidrule(lr){2-5}\cmidrule(lr){6-9}\cmidrule(lr){10-13}
    
    & \cellcolor{proactivepurple!22}Timing 
    & \cellcolor{proactivepurple!22}Under. 
    & \cellcolor{proactivepurple!22}Repeat. 
    & \cellcolor{proactivepurple!22}Avg. 
    
    & \cellcolor{proactiveblue!18}Cancel 
    & \cellcolor{proactiveblue!18}Modify 
    & \cellcolor{proactiveblue!18}Multi 
    & \cellcolor{proactiveblue!18}Avg. 
    
    & \cellcolor{proactiveindigo!22}R2P 
    & \cellcolor{proactiveindigo!22}RuP 
    & \cellcolor{proactiveindigo!22}RaP 
    & \cellcolor{proactiveindigo!22}Avg. \\ 
    \midrule
    Human Level & 98.00 & 94.00 & 92.00 & 94.67 & 98.00 & 92.00 & 98.00 & 96.00 & 96.00 & 96.00 & 98.00 & 96.67 \\
    \midrule
    \multicolumn{13}{@{}l@{}}{\textit{Proprietary Models-Offline}} \\
    Gemini 3 Pro & 43.49 & 24.90 & \cellcolor{red!20}18.40 & 28.93 & 19.44 & 21.84 & \cellcolor{red!20}19.39 & 20.22 & 24.76 & 17.68 & 26.69 & 23.04 \\
    Gemini 2.5 Pro & 45.36 & 25.26 & 13.21 & 27.94 & 24.54 & 20.81 & 16.78 & 20.71 & 21.17 & 17.54 & 28.24 & 22.32 \\
    GPT-5.4 & \cellcolor{red!20}54.64 & 23.52 & 8.96 & 29.04 & \cellcolor{red!20}38.43 & 24.50 & 7.57 & \cellcolor{red!20}23.50 & 27.04 & 15.06 & 29.28 & 23.79 \\
    GPT-4o & 50.59 & 25.99 & 13.68 & \cellcolor{red!20}30.09 & 11.57 & \cellcolor{red!20}31.88 & 14.18 & 19.21 & 27.04 & 17.68 & 26.17 & 23.63 \\
    \midrule
    \multicolumn{13}{@{}l@{}}{\textit{Open-source Models-Offline}} \\
    LLaVA-OneVision-7B & 12.14 & 8.66 & 0.00 & 6.93 & 1.39 & 1.01 & 0.47 & 0.96 & 3.91 & 0.00 & 3.89 & 2.60 \\
    InternVL3-8B & 32.51 & 22.31 & 3.77 & 19.53 & 6.94 & 6.04 & 3.55 & 5.51 & 7.17 & 0.00 & 9.33 & 5.50 \\
    Qwen3-VL-8B & 43.12 & 24.51 & 4.25 & 23.96 & 1.39 & 17.45 & 9.22 & 9.35 & 24.43 & 19.75 & 23.05 & 22.41 \\
    Qwen3.5-Plus & 35.25 & 24.35 & 7.89 & 22.50 & 12.96 & 31.54 & 10.87 & 18.46 & \cellcolor{red!20}32.25 & \cellcolor{red!20}22.01 & 30.57 & \cellcolor{red!20}28.28 \\
    GLM-4.6V & 50.43 & \cellcolor{red!20}27.59 & 10.85 & 29.62 & 15.28 & 24.16 & 9.69 & 16.38 & 22.15 & 6.77 & \cellcolor{red!20}39.37 & 22.76 \\
    \midrule
    \multicolumn{13}{@{}l@{}}{\textit{Open-source Models-Online}} \\
    VideoLLM-online-8B & 14.63 & 0.14 & 1.82 & 5.53 & 7.87 & 1.68 & 6.62 & 5.39 & 0.00 & 1.10 & 1.55 & 0.88 \\
    Dispider & 18.72 & 0.65 & 1.82 & 7.06 & - & - & - & - & - & - & - & - \\
    Flash-VStream-7B & 5.91 & 0.00 & 0.00 & 1.97 & 2.31 & 0.00 & 0.00 & 0.77 & 0.00 & 0.00 & 0.78 & 0.26 \\
    \bottomrule
    \end{tabularx}
\end{table*}

\subsection{Results on IPIBench}
\label{sec:results}
We evaluate a broad range of models on IPIBench, including proprietary offline MLLMs such as Gemini 3 Pro\cite{team2023gemini}, Gemini 2.5 Pro\cite{comanici2025gemini}, GPT-5.4\cite{singh2025openai}, and GPT-4o\cite{hurst2024gpt}, open-source offline MLLMs such as LLaVA-OneVision-7B\cite{li2024llava}, InternVL3-8B\cite{zhu2025internvl3}, Qwen3-VL-8B\cite{bai2025qwen3}, Qwen3.5-Plus\cite{team2026qwen3}, and GLM-4.6V\cite{hong2025glm}, as well as open-source online streaming models including VideoLLM-online-8B\cite{chen2024videollm}, Dispider\cite{qian2025dispiderenablingvideollms}, and Flash-VStream-7B\cite{zhang2024flash}. We additionally conduct human evaluation for comparison. The overall results are reported in Tab.~\ref{tab:main}.

Overall, proprietary models consistently outperform open-source models on both \emph{Proactive Monitoring} and \emph{Proactive Task Management} tasks. On \emph{Interleaved Reactive--Proactive Requests}, proprietary models still achieve generally stronger performance overall, while the latest open-source model Qwen3.5-Plus attains the best performance among all evaluated models. 
Although existing offline MLLMs are primarily trained for reactive VQA tasks, they still exhibit non-trivial proactive interactive capabilities through prompt-based adaptation, and even outperform several carefully designed online streaming models. In contrast, models trained for specific streaming video tasks perform poorly on IPIBench. Compared with single-turn \emph{Proactive Monitoring} tasks, \emph{Proactive Task Management} and \emph{Interleaved Reactive--Proactive Requests} are substantially more challenging for all model families. Finally, despite the strong performance of recent proprietary MLLMs, even the best-performing model still remains far below human-level performance, indicating that interactive proactive intelligence under streaming settings is still far from solved.

    


\begin{wraptable}[15]{R}{0.5\textwidth}
    \centering
    \scriptsize
    \vspace{-2.5em}
    \caption{\textbf{Failure analysis on IPIBench.} We report the distribution (\%) of early and late triggers among incorrect cases in \textit{Proactive Timing}, and the performance improvement on the \textit{Reactive-under-Proactive Interaction} (RuP) task when an explicit reminder instruction (with Ins.) is provided.}
    \label{tab:error}

    \setlength{\tabcolsep}{1.1pt}
    \renewcommand{\arraystretch}{1.1}

    \begin{tabular*}{0.98\linewidth}{@{\extracolsep{\fill}} l c c c c @{}}
    \toprule
    \multirow{2}{*}{Model} &
    \multicolumn{2}{>{\columncolor{proactivepurple!55}}c}{Timing Error} &
    \multicolumn{2}{>{\columncolor{proactiveindigo!55}}c@{}}{RuP} \\
    \cmidrule(lr){2-3} \cmidrule(lr){4-5}
    & \cellcolor{proactivepurple!22}Early
    & \cellcolor{proactivepurple!22}Late
    & \cellcolor{proactiveindigo!22}Base
    & \cellcolor{proactiveindigo!22}with Ins. \\
    \midrule
    Gemini 3 Pro         & 92.66 & 7.34  & 17.68 & 19.20 \\
    GPT-5.4              & 51.30 & 48.70 & 15.06 & 19.34 \\
    Qwen3.5-Plus         & 93.42 & 6.58  & 22.01 & 26.80 \\
    Qwen3-VL-8B          & 31.48 & 68.52 & 19.75 & 20.30 \\
    InternVL3-8B         & 29.82 & 70.18 & 0.00  & 0.28  \\
    LLaVA-OneVision-7B   & 8.39  & 91.61 & 0.00  & 0.97  \\
    \bottomrule
    \end{tabular*}
\end{wraptable}
\subsection{Failure Analysis}
\label{sec:failure}
To better understand why existing MLLMs perform substantially worse on IPIBench than on conventional VQA benchmarks, we further analyze model failures on two representative tasks. The results are summarized in Tab.~\ref{tab:error}.
We first investigate failure patterns on \emph{Proactive Timing}, which serves as the foundational task of IPIBench. Interestingly, we observe substantially different temporal triggering behaviors across model families.  Larger proprietary and open-source models tend to aggressively activate proactive responses even with incomplete visual evidence, whereas smaller models often fail to react promptly due to weaker visual grounding capability. This suggests that existing MLLMs lack stable temporal triggering policies under continuous streams. 


We further analyze \emph{Reactive-under-Proactive Interaction} task to better understand the difficulty of multi-turn interactive proactive behavior. In this setting, a reactive query is inserted before the proactive task is triggered, potentially interrupting the ongoing proactive monitoring process. To analyze this effect, we additionally append a reminder instruction after the reactive interaction, explicitly asking the model whether it should generate a proactive response based on the historical context. As shown in Tab.~\ref{tab:error}, all models consistently achieve improved performance after introducing this additional instruction. This indicates that existing MLLMs struggle to consistently maintain proactive objectives throughout multi-turn interactions without explicit guidance. Overall, these failure analyses demonstrate that IPIBench introduces challenges beyond conventional streaming VQA settings, requiring models not only to understand visual content, but also to perform temporally stable proactive triggering and long-term interaction coordination under continuous video streams.

\begin{table*}[t]
    \centering
    \scriptsize
    \caption{\textbf{Effectiveness of the IPI-Agent framework.} We evaluate our proposed agentic framework using four representative base MLLMs. The \textcolor{green!70!black}{green} numbers denote the gain compared to the original base models (results in Tab.~\ref{tab:main}).}
    \label{tab:agent}
    \setlength{\tabcolsep}{1.9pt} 
    \renewcommand{\arraystretch}{1.8} 
    
    \newcommand{\inc}[2]{\shortstack{#1\\ \tiny\textcolor{green!70!black}{+#2}}}
    \newcommand{\dec}[2]{\shortstack{#1\\ \tiny\textcolor{red!80!black}{-#2}}}

    \begin{tabularx}{0.98\linewidth}{@{} L{2.8} *{12}{C{0.75}} @{}}
    \toprule
    \multirow{2}{*}{Model} 
    & \multicolumn{4}{c}{\cellcolor{proactivepurple!55}Proactive Monitoring} 
    & \multicolumn{4}{c}{\cellcolor{proactiveblue!45}Proactive Task Management} 
    & \multicolumn{4}{c@{}}{\cellcolor{proactiveindigo!55}Interleaved Reactive--Proactive} \\
    \cmidrule(lr){2-5}\cmidrule(lr){6-9}\cmidrule(lr){10-13}
    
    & \cellcolor{proactivepurple!22}Timing 
    & \cellcolor{proactivepurple!22}Under. 
    & \cellcolor{proactivepurple!22}Repeat. 
    & \cellcolor{proactivepurple!22}Avg. 
    
    & \cellcolor{proactiveblue!18}Cancel 
    & \cellcolor{proactiveblue!18}Modify 
    & \cellcolor{proactiveblue!18}Multi 
    & \cellcolor{proactiveblue!18}Avg. 
    
    & \cellcolor{proactiveindigo!22}R2P 
    & \cellcolor{proactiveindigo!22}RuP 
    & \cellcolor{proactiveindigo!22}RaP 
    & \multicolumn{1}{c@{}}{\cellcolor{proactiveindigo!22}Avg.} \\ 
    \midrule
    
    IPI-Agent (Gemini 3 Pro) 
    & \inc{56.27}{12.78} & \inc{35.67}{10.77} & \inc{28.30}{9.90} & \inc{40.08}{11.15} 
    & \inc{51.85}{32.41} & \inc{35.23}{13.39} & \inc{29.08}{9.69} & \inc{38.72}{18.50} 
    & \inc{30.62}{5.86}  & \inc{22.38}{4.70}  & \inc{29.53}{2.84} & \inc{27.51}{4.47} \\
    
    IPI-Agent (GPT-5.4) 
    & \inc{57.20}{2.56} & \inc{24.30}{0.78} & \inc{10.85}{1.89} & \inc{30.78}{1.74} 
    & \inc{48.61}{10.18} & \inc{24.83}{0.33} & \inc{14.66}{7.09} & \inc{29.37}{5.87} 
    & \inc{28.01}{0.97}  & \inc{19.34}{4.28}  & \inc{33.42}{4.14} & \inc{26.92}{3.13} \\
    
    IPI-Agent (Qwen3.5-Plus)
    & \inc{52.80}{17.55} & \inc{33.18}{8.83} & \inc{18.42}{10.53} & \inc{34.80}{12.30} 
    & \inc{52.78}{39.82} & \inc{33.22}{1.68} & \inc{25.53}{14.66} & \inc{37.18}{18.72} 
    & \inc{32.25}{0.00}   & \inc{22.63}{0.62}  & \inc{35.95}{5.38} & \inc{30.28}{2.00} \\

    IPI-Agent (Qwen3-VL-8B)
    & \inc{46.62}{3.50} & \inc{25.37}{0.86} & \inc{15.09}{10.84} & \inc{29.03}{5.07} 
    & \inc{44.91}{43.52} & \inc{23.49}{6.04} & \inc{11.82}{2.60} & \inc{26.74}{17.39} 
    & \inc{27.36}{2.93}  & \inc{23.62}{3.87}  & \inc{30.57}{7.52} & \inc{27.18}{4.77} \\
    
    \bottomrule
    \end{tabularx}
\end{table*}

\subsection{Results of IPI-Agent}
\label{sec:agent_results}
We further adopt four representative MLLMs, including Gemini 3 Pro, GPT-5.4, Qwen3.5-Plus, and Qwen3-VL-8B, as the base models of IPI-Agent to evaluate the effectiveness of the proposed agentic framework. In all experiments, we use Qwen3-VL-Embedding-2B\cite{li2026qwen3} as the embedding model for the Gating Tool. The results are summarized in Tab.~\ref{tab:agent}. Overall, IPI-Agent consistently improves performance across all benchmark aspects and model families, demonstrating the effectiveness of the proposed interaction-control policy and Gating Tool under interactive streaming settings.
In particular, IPI-Agent achieves the largest improvements on \emph{Proactive Task Management}, which requires models to correctly handle various user management instructions under continuous streams. Overall, the results demonstrate that IPI-Agent effectively addresses the two major limitations revealed by IPIBench, namely unstable proactive triggering and weak multi-turn interaction coordination, providing a simple yet effective framework for interactive proactive tasks under streaming settings.

\begin{table*}[t]
    \centering
    \scriptsize
    \caption{\textbf{Ablation studies on IPI-Agent components.} We evaluate different variants of IPI-Agent to investigate the contribution of the interaction-control policy and the temporal-gating mechanism. \textcolor{red!80!black}{Red} numbers indicate performance degradation compared to the full IPI-Agent (based on Qwen3-VL-8B).}
    \label{tab:ablation}
    \setlength{\tabcolsep}{1.9pt} 
    \renewcommand{\arraystretch}{1.4}

    \newcommand{\inc}[1]{\textcolor{green!70!black}{+#1}}
    \newcommand{\dec}[1]{\textcolor{red!80!black}{-#1}}

    \begin{tabularx}{0.98\linewidth}{@{} L{2.8} *{12}{C{0.75}} @{}}
    \toprule
    \multirow{2}{*}{Variant}
    & \multicolumn{4}{c}{\cellcolor{proactivepurple!55}Proactive Monitoring}
    & \multicolumn{4}{c}{\cellcolor{proactiveblue!45}Proactive Task Management}
    & \multicolumn{4}{c@{}}{\cellcolor{proactiveindigo!55}Interleaved Reactive--Proactive} \\
    \cmidrule(lr){2-5}\cmidrule(lr){6-9}\cmidrule(lr){10-13}

    & \cellcolor{proactivepurple!22}Timing
    & \cellcolor{proactivepurple!22}Under.
    & \cellcolor{proactivepurple!22}Repeat.
    & \cellcolor{proactivepurple!22}Avg.

    & \cellcolor{proactiveblue!18}Cancel
    & \cellcolor{proactiveblue!18}Modify
    & \cellcolor{proactiveblue!18}Multi
    & \cellcolor{proactiveblue!18}Avg.

    & \cellcolor{proactiveindigo!22}R2P
    & \cellcolor{proactiveindigo!22}RuP
    & \cellcolor{proactiveindigo!22}RaP
    & \cellcolor{proactiveindigo!22}Avg. \\
    \midrule 
    IPI-Agent (Qwen3-VL-8B) & 46.62 & 25.37 & 15.09 & 29.03 & 44.91 & 23.49 & 11.82 & 26.74 & 27.36 & 23.62 & 30.57 & 27.18 \\

    \midrule

    w/o Interaction Control
    & \textcolor{red!80!black}{0.00} & \textcolor{red!80!black}{0.00} & \textcolor{red!80!black}{0.00} & \textcolor{red!80!black}{0.00}
    & \dec{40.95} & \dec{4.70} & \dec{0.24} & \dec{15.54}
    & \dec{1.81} & \dec{0.55} & \dec{3.11} & \dec{1.82} \\

    w/o Temporal Gating
    & \dec{3.50} & \dec{0.86} & \dec{10.84} & \dec{5.07}
    & \dec{6.95} & \dec{5.37} & \dec{1.65} & \dec{4.66}
    & \dec{0.32} & \dec{2.21} & \dec{3.63} & \dec{2.05} \\

    \bottomrule
    \end{tabularx}
\end{table*}

\subsection{Ablation Study}
\label{sec:ablation}

We further conduct ablation studies to analyze the effectiveness of different components in IPI-Agent. The results are summarized in Tab.~\ref{tab:ablation}.
We first evaluate IPI-Agent without the interaction-control policy while keeping the temporal-gating mechanism unchanged. This variant causes the largest performance drop on \emph{Proactive Task Management}, highlighting the importance of explicit interaction coordination for handling user management instructions under continuous streams. We further remove the temporal-gating mechanism from IPI-Agent. Compared with the full framework, performance consistently degrades across all task categories, especially on \emph{Proactive Monitoring}, demonstrating the effectiveness of temporal gating for stabilizing proactive triggering. 

Finally, we additionally investigate whether proactive triggering can be achieved using only embedding-based semantic similarity without the proposed agentic framework. Using Qwen3-VL-Embedding-2B and Qwen3-VL-Embedding-8B as pure embedding-based trigger models, the performance on \emph{Proactive Timing} drops by 13.41 and 10.36 points, respectively, while \emph{Repeated Proactiveness} drops by 15.09 points for both models. These results suggest that semantic similarity alone is insufficient for reliable proactive triggering, and effective interactive proactive behavior still requires higher-level reasoning capabilities.

%% file: sec/6_conclusion.tex
\section{Conclusion}
\label{sec:conclusion}
In this work, we introduced \textbf{IPIBench}, the first benchmark for evaluating interactive proactive intelligence of MLLMs under streaming video settings. Unlike prior work that studies reactive or proactive interactions separately, IPIBench systematically covers proactive monitoring, proactive task management, and interleaved reactive--proactive requests in dynamic multi-turn environments. Our evaluations reveal that existing MLLMs struggle with stable proactive triggering and coordinated reactive--proactive interaction under continuous streams. To address these limitations, we further propose \textbf{IPI-Agent}, a training-free agentic framework with an interaction-control policy and a temporal-gating mechanism. Experimental results show that IPI-Agent consistently improves existing MLLMs across diverse tasks and interaction settings. We hope our benchmark and framework can facilitate future research on interactive proactive multimodal intelligence.